\documentclass{article}

\usepackage{microtype}
\usepackage{graphicx}
\usepackage{subfigure}
\usepackage{booktabs} 
\usepackage{array} 
\usepackage{hyperref}
\usepackage{afterpage}
\usepackage[utf8]{inputenc}
\usepackage{newunicodechar}
\newunicodechar{✱}{*}

\usepackage[accepted]{icml2025}

\usepackage{amsmath}
\usepackage{amssymb}
\usepackage{mathtools}
\usepackage{amsthm}
\usepackage{placeins}
\usepackage{float}
\usepackage{placeins}
\usepackage[capitalize,noabbrev]{cleveref}

\theoremstyle{plain}

\theoremstyle{definition}

\theoremstyle{remark}

\usepackage[textsize=tiny]{todonotes}

\icmltitlerunning{ALCo-FM}

\begin{document}

\twocolumn[
\icmltitle{
ALCo-FM: Adaptive Long-Context Foundation Model for Accident Prediction
}



\icmlsetsymbol{equal}{*}

\begin{icmlauthorlist}
\icmlauthor{Pinaki Prasad Guha Neogi}{yyy}
\icmlauthor{Ahmad Mohammadshirazi}{yyy}
\icmlauthor{Rajiv Ramnath}{yyy}
\end{icmlauthorlist}

\icmlaffiliation{yyy}{Department of Computer Science and Engineering, Ohio State University, Ohio, US}

\icmlcorrespondingauthor{Pinaki Prasad Guha Neogi}{guhaneogi.2@osu.edu}

\icmlkeywords{Traffic Accident Prediction, Multi-Modal Framework, Machine Learning, Vision Transformers (ViTs), Graph Neural Networks (GNNs), Long-Context, Foundation Model}

\vskip 0.3in
]



\printAffiliationsAndNotice{}  


\begin{abstract}

Traffic accidents are rare, yet high-impact events that require long-context multimodal reasoning for accurate risk forecasting. In this paper, we introduce \textbf{ALCo-FM}, a unified adaptive long-context foundation model that computes a volatility pre-score to dynamically select context windows for input data and encodes and fuses these multimodal data via shallow cross attention. Following a local GAT layer and a BigBird‐style sparse global transformer over H3 hexagonal grids, coupled with Monte Carlo dropout for confidence, the model yields superior, well‐calibrated predictions. Trained on data from 15 U.S. cities with a class-weighted loss to counter label imbalance, and fine-tuned with minimal data on held-out cities, ALCo-FM achieves 0.94 accuracy, 0.92 F1, and an ECE of 0.04—outperforming 20+ state-of-the-art baselines in large-scale urban risk prediction. Code and dataset are available at: 
https://github.com/PinakiPrasad12/ALCo-FM


\end{abstract}

\section{Introduction}
\label{sec:intro}

\begin{figure*}[h!]  
    \centering
    \includegraphics[width=1\linewidth]{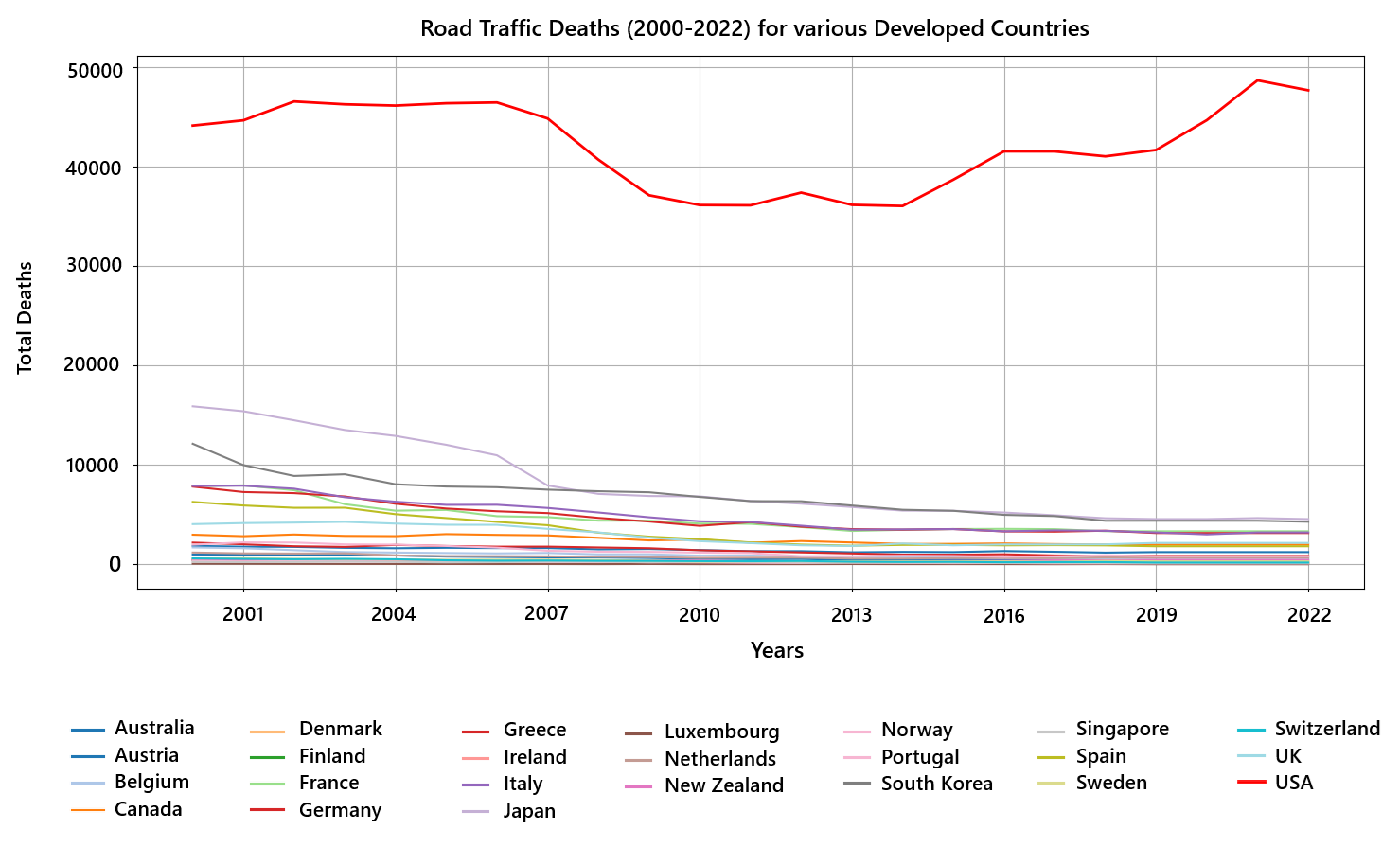}
    \caption{Deaths due to road accidents in various developed countries (2000--2022).}
    \label{fig:road_deaths}
\end{figure*}

\begin{figure*}[h!]  
    \centering
    \includegraphics[width=1\linewidth]{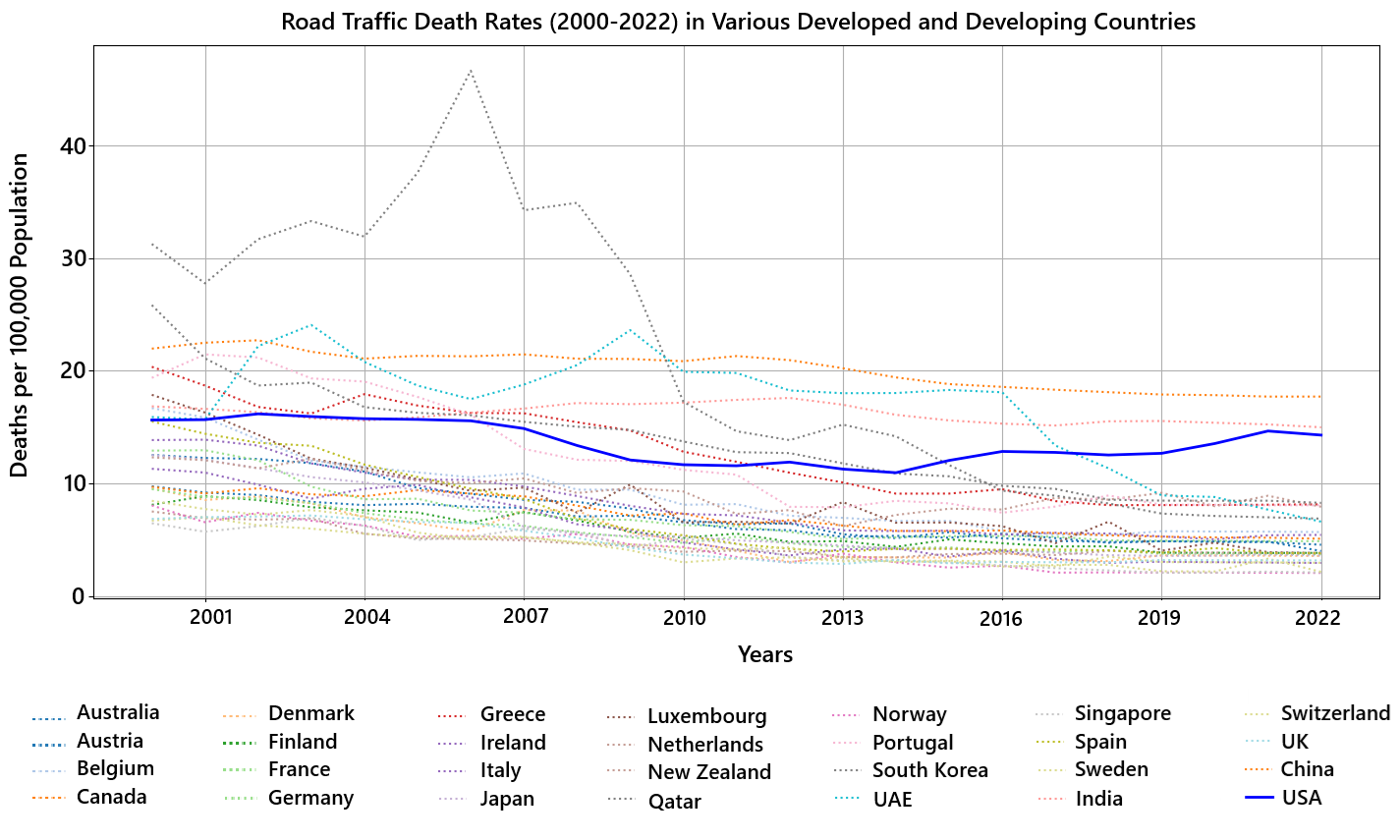}
    \caption{Road accident fatality rates (per 100,000 population) in various countries (2000--2022).}
    \label{fig:road_death_rates}
\end{figure*}

Mitigating road accidents stands out as one of the most pressing challenges in modern transportation, claiming numerous lives annually and imposing significant economic burdens worldwide \cite{khan2020evaluating, yu2021deep, berhanu2023examining}.
According to the latest data from the World Health Organization (WHO)~\cite{who2024} and the Insurance Institute for Highway Safety (IIHS)~\cite{iihs2024}, global efforts to reduce traffic fatalities—through policy reforms, technological advancements, and public safety initiatives—have led to measurable improvements in many developed countries. As shown in Figure~\ref{fig:road_deaths}, annual road death counts from 2000 to 2022 exhibit clear downward trends in several first-world nations, largely attributed to improved vehicle safety, stricter enforcement, and awareness campaigns.
However, the United States presents a notable exception. Both the total number of road fatalities and the per capita fatality rate (Figure~\ref{fig:road_death_rates}) remain significantly higher than in peer nations and display erratic or stagnating trends. Even when compared to high-population countries such as India and China, U.S.\ fatality rates remain elevated, with a relatively flat or rising trajectory—contrasting the slow but consistent declines observed in those countries.
Interestingly, countries with rapidly transformed economies like Qatar and the UAE, which once reported much higher fatality rates than the U.S., have experienced sharp declines in recent years and now report rates lower than those of the United States. This divergence underscores the urgency of understanding the persistent and complex nature of traffic fatalities in the U.S., motivating the need for more robust, multimodal predictive frameworks.

Existing forecasting methods—from ensemble and transformer-based time-series models to Vision Transformers on map tiles and spatio-temporal GNNs—typically (1) assume a short, fixed history window, (2) operate on a single data modality, or (3) omit uncertainty calibration, and most critically, (4) fail to model long-range context jointly across modalities. These gaps limit their effectiveness in real-world, large-scale deployments.


To address these challenges, we introduce the \textbf{Adaptive Long‐Context Foundation Model (ALCo-FM)}, whose key innovations are:

\vspace{-2mm}

\begin{itemize}
  \item \textbf{Volatility‐Driven Context Selection}  
   We compute a lightweight pre‐score from each 1h ContiFormer~\cite{chen2024contiformer} + T2T-ViT~\cite{Yuan_2021_ICCV} embedding to gate between 1h, 3h, or 6h look-back windows, ensuring more history is used when and where it matters.

  \item \textbf{Unified Dual‐Transformer Encoding \& Fusion}  
   Numerical time‐series and map imagery are encoded in parallel by a ContiFormer
   and a T2T-ViT
   , then seamlessly fused via shallow cross‐attention to capture rich temporal–spatial interactions.

  \item \textbf{Scalable Hybrid Attention on H3 Grids}  
   We first propagate local neighborhood information with a GAT layer, then apply a BigBird‐style sparse global transformer for efficient, city-wide context aggregation across all H3 cells~\cite{uber_h3}.

  \item \textbf{Foundation-Scale Calibration \& Generalization}  
   A 2-layer MLP head is calibrated with Monte Carlo dropout for reliable uncertainty estimates, and ALCo-FM is pretrained on 15 U.S. cities with a class‐weighted loss, then fine-tuned on new regions using minimal data.
\end{itemize}

\vspace{-2mm}

These advances combine to deliver robust, long-context multimodal accident-risk forecasting at urban scale.  
Evaluated on 1,771 regions across 15 U.S.\ cities—and, after minimal fine‐tuning, on 3 held‐out cities—our model sets new benchmarks in accuracy, F1, and calibration, demonstrating both long‐context capability and foundation‐model versatility.


\section{Related Work}
\label{sec:related}

Existing approaches to traffic accident prediction focus primarily on three main methodological paradigms—numerical and time-series modeling, vision-based learning, and spatio-temporal GNNs. Here, we summarize the notable developments within each of these methodological categories:

\textbf{(1) Numerical and Time-Series Modeling:} Numerical features such as road type, traffic volume, and weather conditions play a central role in accident prediction. Traditional methods rely on ensemble classifiers like Random Forest \cite{Breiman2001}, XGBoost \cite{Chen_2016}, as well as hybrid models for time-series forecasting~\cite{ahmed2023study, koc2022accident}. More recently, transformer-based frameworks have shown promise in handling irregular temporal structures. Models like \textbf{ContiFormer}~\cite{chen2024contiformer}, \textbf{MSCA}~\cite{hammad2024multi}, and \textbf{CST}~\cite{fonseca2023continuousspatiotemporaltransformers} have demonstrated robust sequence modeling capabilities, with potential applicability to accident prediction, where data sparsity and irregular incident patterns pose challenges.

\textbf{(2) Vision-Based Approaches:} Visual data offers rich contextual cues for accident risk assessment. CNN-based models have been previously used to extract road features, including lane structures and local infrastructure~\cite{ahmed2023study}. Recent advancements in \textbf{Vision Transformers (ViTs)} enable high-level spatial reasoning through self-attention mechanisms. While not specifically designed for accident prediction, models like \textbf{ViT-CoMer}~\cite{xia2024vit} and \textbf{T2T-ViT}~\cite{Yuan_2021_ICCV} highlight the potential of leveraging map and satellite imagery for traffic safety modeling.

\textbf{(3) Spatio-Temporal Graph Neural Networks:} GNN-based approaches model both spatial correlations and temporal dependencies in accident prediction. Early works such as \textbf{DCRNN}~\cite{li2018diffusionconvolutionalrecurrentneural} and \textbf{STGCN}~\cite{ijcai2018p505} established the foundation for graph-based traffic forecasting. Subsequent studies introduced attention mechanisms~\cite{guo2019attention}, adaptive adjacency structures-Graph WaveNet (\textbf{GWN})~\cite{wu2019graphwavenetdeepspatialtemporal}, and differential equation-based modeling~\cite{Fang_2021} to improve long-range dependencies. Automated architecture search further enhances GNN-based forecasting through methods like \textbf{AutoSTG}~\cite{10.1145/3442381.3449816} and \textbf{Auto-DSTSG}~\cite{jin2022automated}.

For accident prediction specifically, recent GNN-based models address data sparsity and regional heterogeneity by incorporating multi-resolution representations. Approaches such as \textbf{DSTGCN}~\cite{yu2021deep}, \textbf{DGCRN}~\cite{li2023dynamic}, and \textbf{SST-GCN}~\cite{kim2024sst} integrate external features like weather and POIs to enhance predictive accuracy. More advanced architectures, including \textbf{STSGCN}~\cite{Song_Lin_Guo_Wan_2020}, \textbf{STFGNN}~\cite{Li_Zhu_2021}, \textbf{STGODE}~\cite{Fang_2021}, \textbf{VSTGCN}~\cite{gan2024novel}, \textbf{UTAASTRL}~\cite{bao2020uncertainty}, and \textbf{FC-STGNN}~\cite{wang2024fully}, employ fusion layers, dynamic adjacency matrices, and self-supervised mechanisms to handle accident rarity and distribution skew. Other recent GNN-based efforts further explore interpretability and dual-branch reasoning, including \textbf{ITAP}~\cite{10693871} for self-explanatory accident modeling and \textbf{DualRisk}~\cite{10716568} for handling occurrence-severity asynchronicity in sparse accident data.

Recent efforts further integrate GNNs with transformer-driven or vision-enabled components to unify multimodal data. Models like \textbf{GTC}~\cite{SUN2025106645} leverage co-contrastive learning between GNNs and transformers, while \textbf{MST-GAT}~\cite{DING2023527} combines multimodal graph attention to enrich spatial-temporal learning. Additionally, models such as \textbf{Het-ConvLSTM}~\cite{yuan2018hetero} introduce spatial graph features into ConvLSTM models, while \textbf{SST-DHL}~\cite{cui2024advancing} incorporates dynamic hierarchical learning for improved forecasting. Further, \textbf{MVMT-STN}~\cite{wang2021traffic} and \textbf{GLST-TARP}~\cite{alhaek2025encoding} explore multi-scale spatio-temporal networks and global-local traffic accident risk prediction, respectively.
Although initially developed for broader applications, these techniques hold strong potential for traffic accident risk assessment when adapted to domain-specific constraints. 
Recent models such as \textbf{UTARPR}~\cite{Chen_2024} also emphasize hierarchical regional semantics for traffic risk mapping, but they don't integrate visual or tabular modalities.

While several recent works~\cite{wu2024accidentgptlargemultimodalfoundation, s23229225, 10489808, wang2021riskpredictiontrafficaccidents} have explored multi-modal accident analysis, they fall short of integrating all three critical modalities—structured numerical time-series data, spatial imagery, and graph-based relational context. For instance,~\cite{wang2021riskpredictiontrafficaccidents} combines visual and temporal cues using compact dynamic features but lacks explicit graph modeling. ~\cite{10489808} uses knowledge graphs for interpretability but omits image-based spatial reasoning. Others~\cite{s23229225, wu2024accidentgptlargemultimodalfoundation} focus on language or video modalities for broader tasks like intervention or reconstruction, diverging from our goal of predictive risk modeling.

By contrast, our proposed framework unites uncertainty‐driven context window selection, cross‐modal fusion of time‐series and imagery, and a hybrid local–global attention mechanism, providing scalable long‐range reasoning with calibrated risk estimation in a single end‐to‐end model.


\section{Dataset Description}
\label{sec:data}

We construct a high-resolution, multi-source dataset tailored for adaptive long-context accident prediction.

\subsection{Data Preparation}
\label{subsec:data_preparation}

We aggregate four heterogeneous data sources on an H3 hexagonal grid at resolution \(R=7\)~\cite{uber_h3}. First, we incorporate traffic events from the nationwide dataset of Moosavi et al.~\cite{moosavi2019countrywide}, which spans 49 U.S.\ states and provides precise timestamps, GPS locations, and contextual details per incident (Appendix~\ref{app:App_Btraffic}). Second, we enrich each cell with demographic attributes—150 socio‐economic variables such as income, population density, and age distribution—sourced for 45,000 ZIP codes from the U.S.\ ZIP Code database~\cite{us_zipcodes} (Appendix~\ref{app:App_Bdemo}). Third, we align hourly meteorological observations (2016–2023) from the Iowa Environmental Mesonet’s ASOS network~\cite{iem_asos}, including temperature, precipitation, and wind speed, to each accident record (Appendix~\ref{app:App_Bweather}). Finally, for spatial context we download \(256\times256\) map tiles from OpenStreetMap~\cite{OpenStreetMap} centered at each hexagon’s centroid (Appendices~\ref{app:App_Bimg}, \ref{app:App_C3}). Together, these inputs form a rich, multi‐modal foundation for our adaptive long‐context modeling.

These data sources, combined with rigorous preprocessing and advanced feature engineering, allow us to model the intricate dependencies influencing accident risk. Further dataset details and feature descriptions are provided in Appendix~\ref{apn:appen_C}. Also, a detailed statistical and structural analysis of our dataset is provided in Appendix~\ref{apn:appen_dataset_analysis}, which highlights its high dimensionality, severe class imbalance, weak feature correlations, and nonlinear separability—underscoring its real-world complexity and the need for advanced modeling techniques.
The final dataset for our experiment consists of 1,771 regions across 15 U.S. cities as shown in Table~\ref{tab:final_nodes}. The process of selecting these cities has been elaborately explained in Appendix \ref{app:App_C4}.

\begin{table}[ht!]
\centering
\caption{Final Number of Nodes in Each Selected City}
\label{tab:final_nodes}
\setlength{\tabcolsep}{15pt}
  \renewcommand{\arraystretch}{1}
\begin{tabular}{l|l|c}
\toprule
\textbf{City} & \textbf{State} & \textbf{No. of Nodes} \\
\midrule
Atlanta & GA & 88 \\
Austin & TX & 139 \\
Baton Rouge & LA & 64 \\
Charlotte & NC & 183 \\
Dallas & TX & 140 \\
Houston & TX & 246 \\
Los Angeles & CA & 85 \\
Miami & FL & 152 \\
Minneapolis & MN & 84 \\
Nashville & TN & 93 \\
Orlando & FL & 150 \\
Phoenix & AZ & 106 \\
Raleigh & NC & 97 \\
Sacramento & CA & 74 \\
San Diego & CA & 70 \\
\midrule
\textbf{Total} & - & \textbf{1,771} \\
\bottomrule
\end{tabular}
\end{table}

\vspace{-3mm}

\paragraph{Long-context.}  
Although in our experiment we consider only 1-6 hours window (\ref{sec:methodology-4_1}), each hour yields a high-dimensional, multimodal embedding: up to hundreds of numeric features and \(P\) visual patch tokens per cell. A 6h window therefore produces on the order of \(6T + 6P\) tokens per node, and we must reason over all 1,771 cells simultaneously. This naturally gives rise to a “long-context” modeling problem, motivating our adaptive windowing, cross-modal fusion, and sparse global attention mechanisms (Section~\ref{sec:methodology}).

\subsection{Data Preprocessing}
\label{subsec:data_preprocessing}

All sources are first merged on the H3 cell index and timestamp and then aligned into uniform \textbf{1-hour} atomic windows. We enrich each atomic window with temporal indicators (rush hour, part of day, U.S.\ holidays) and simple geographic summaries (neighbor counts, road density), impute any remaining gaps via FAISS k-NN~\cite{johnson2019billion}
(Refer Appendix~\ref{apn:appen_A} for full preprocessing pipeline and rationale). These 1-hour windows are the input units that get dynamically grouped into adaptive long-context spans in the model (Section~\ref{sec:methodology-4_1}).


\section{Methodology}
\label{sec:methodology}

In this section, we step through our unified adaptive long‐context framework for accident prediction, which seamlessly blends spatial, temporal, and visual information within a single model. Detailed descriptions of each component presented in Sections \ref{sec:methodology-4_1}–\ref{sec:methodology-4_5}.

\subsection{H3 Grid Mapping \& Adaptive Long-Context Dual Encoding}
\label{sec:methodology-4_1}

We tessellate the target area into H3 hexagons, yielding nodes \(v_i\). For each \(v_i\), we aggregate traffic, weather, and demographic records over a one-hour atomic window and encode them via paired encoders: 

\vspace{-2mm}

\paragraph{Numerical Encoder (ContiFormer).}  
The 1-hour numerical time series data at node \(v_i\) is processed by a continuous-time state-space transformer (ContiFormer), producing \(T\) tokens $X_{num}\in\mathbb R^{T\times d}$ that capture temporal dynamics and serve both for uncertainty scoring and downstream fusion.

\vspace{-2mm}

\paragraph{Visual Encoder (T2T-ViT).}  
Concurrently, we crop a \(256\times256\) map tile around \(v_i\) and tokenize it with a Token-to-Token Vision Transformer into \(P\) patch tokens $X_{vis}\in\mathbb R^{P\times d}$, encoding spatial context.

The resulting dual embedding \(\bigl[X_{num};X_{vis}\bigr]\) is then used to compute a volatility signal \(u\), which, in turn is used to decide how many hours of history to use. This strategy—similar to ~\cite{graves2017adaptivecomputationtimerecurrent} and ~\cite{sukhbaatar2019adaptiveattentionspantransformers}—ensures that more history is used when volatility is high, and compute is conserved otherwise):
\[
  \sigma_{\mathrm{num}}
  = \tfrac{1}{d}\sum_{j=1}^d\mathrm{std}_t\bigl(X_{\mathrm{num}}[:,j]\bigr)
\]
\[  
  \sigma_{\mathrm{vis}}
  = \tfrac{1}{d}\sum_{j=1}^d\mathrm{std}_p\bigl(X_{\mathrm{vis}}[:,j]\bigr)
\]
\[  
  u = \tfrac12\bigl(\sigma_{\mathrm{num}}+\sigma_{\mathrm{vis}}\bigr)
\]

Here \(\mathrm{std}_t\) is computed across the \(T\) temporal tokens, \(\mathrm{std}_p\) across the \(P\) visual patches.  We then set thresholds \(\tau_{\mathrm{low}},\tau_{\mathrm{high}}\) to the 33rd/67th percentiles of all training‐set \(u\), refining via validation grid‐search to optimize F1 and calibration. And, based on the volatility score \(u\), we choose a look‐back window \(w\) (in hours) as:

\vspace{-2mm}

\[
  w \in \{1,3,6\},\quad
  w=\begin{cases}
    6 & u>\tau_{high},\\
    3 & \tau_{low}\le u\le \tau_{high},\\
    1 & u<\tau_{low}.
  \end{cases}
\]

This adaptive long-context mechanism uses more history when volatility is high, and conserves compute when the signal is stable; during training, we randomly sample \(w\) to build robustness across spans.

\subsection{Cross-Modal Long-Context Fusion}

After determining the optimal historical window length \(w \in \{1,3,6\}\) for each node using the volatility-driven gating strategy (Section~\ref{sec:methodology-4_1}), we retrieve the last \(w\) atomic windows for both modalities—numerical and visual—and re-encode them into long-context token sequences. Specifically, this yields
\[
(X_{\mathrm{num}}, X_{\mathrm{vis}}) \in \mathbb{R}^{wT \times d} \times \mathbb{R}^{wP \times d},
\]
where \(T\) and \(P\) are the number of temporal and visual tokens per hour, and \(d\) is the embedding dimension. These expanded sequences capture extended trends and structural patterns over time.

To integrate the two modalities, we employ a lightweight yet expressive cross-attention mechanism that allows features from one modality to selectively attend to the most relevant signals in the other. Concretely, we use \(L=2\) cross-attention layers, each consisting of a single attention head with hidden dimension \(d_h=128\). The cross-modal attention from numerical to visual tokens is computed as:
\[
  \mathrm{CM}(X_{num},X_{vis}) = \mathcal{S}\bigl(\tfrac{X_{num}W_Q\,(X_{vis}W_K)^{\!\top}}{\sqrt{d_k}}\bigr)\,(X_{vis}W_V)
\]
where \(W_Q, W_K, W_V\) are learned projection matrices, and \(\mathcal{S}(\cdot)\) denotes the softmax function. This attention maps each numerical token to a weighted combination of visual features, enabling the model to incorporate spatial context when interpreting temporal trends. We also compute attention symmetrically in the reverse direction (visual-to-numerical) to allow reciprocal influence.

Each cross-attention layer is wrapped with residual connections and Layer Normalization to preserve gradient flow and stabilize training. After two layers of bidirectional attention, we perform mean-pooling over the long-context sequence to produce two modality-specific fused embeddings per node:
\[
\widetilde{x}_{\mathrm{num}}, \widetilde{x}_{\mathrm{vis}} \in \mathbb{R}^d.
\]

These embeddings are then concatenated and passed to the spatial GAT layer (Section~\ref{sec:local_gat}) for local interaction modeling. This fusion module plays a critical role in capturing the interplay between temporal dynamics and static spatial context, allowing the model to reason over rich, multimodal representations across variable-length windows.

\subsection{Spatio-Temporal Graph Construction \& Local GAT}
\label{sec:local_gat}

To effectively model spatial dependencies and neighborhood-level interactions, we construct a spatial graph \(\mathcal{G} = (\mathcal{V}, \mathcal{E})\) over the \(N\) H3 hexagonal grid cells at each atomic time window. Each node \(v_i \in \mathcal{V}\) represents a specific geographic region, and we define edges \(\mathcal{E}\) by connecting each node to its \(k = 6\) immediate neighbors in the H3 layout, forming a locally connected, fixed-topology graph that is consistent across time slices.

Each node \(v_i\) is associated with two fused modality embeddings—\(\widetilde{x}_{\mathrm{num}}^i\) from the numerical encoder and \(\widetilde{x}_{\mathrm{vis}}^i\) from the visual encoder—both in \(\mathbb{R}^d\). These representations are concatenated to form a joint multimodal feature vector for each node. To enable context-aware reasoning over each node’s neighborhood, we apply a Graph Attention Network (GAT) layer, which learns edge-specific attention weights, allowing the model to assign different importances to different neighbors based on learned relevance.

Formally, for a given node \(v_i\), its updated embedding \(h_i'\) is computed as:
\[
  h_i' 
  = \sigma\left(\sum_{j \in \mathcal{N}(i)} \alpha_{ij} \cdot W_g \left[\widetilde{x}_{\mathrm{num}}^j;\, \widetilde{x}_{\mathrm{vis}}^j\right]\right),
\]
where \(\mathcal{N}(i)\) denotes the set of spatial neighbors of node \(i\), \(W_g \in \mathbb{R}^{2d \times d}\) is a learnable weight matrix for projection, and \(\sigma(\cdot)\) is a nonlinear activation function (we use ReLU in our implementation). The attention coefficient \(\alpha_{ij}\) between node \(i\) and neighbor \(j\) is computed as:

\[
  \alpha_{ij} = \frac{\exp\left(\mathrm{LeakyReLU}\left(a^\top \left[W_g \widetilde{x}^i;\, W_g \widetilde{x}^j\right]\right)\right)}%
    {\sum_{k \in \mathcal{N}(i)} \exp\left(\mathrm{LeakyReLU}\left(a^\top \left[W_g \widetilde{x}^i;\, W_g \widetilde{x}^k\right]\right)\right)},
\]
where \(a \in \mathbb{R}^{2d}\) is a learnable attention vector. This formulation ensures that the influence of each neighbor is adaptively scaled based on its contextual similarity to the target node, as determined by the dot-product in the learned space.

The resulting node embeddings \(\{h_i'\}_{i=1}^N\) are stacked into a matrix:
\[
  Z = 
  \begin{bmatrix}
    (h_1')^\top \\
    (h_2')^\top \\
    \,\vdots \\
    (h_N')^\top
  \end{bmatrix}
  \in \mathbb{R}^{N \times d},
\]
which forms the input to the subsequent global sparse attention layer (Section~\ref{sec:sparse_global}). This locally aggregated feature matrix captures short-range, spatially grounded interactions that are crucial for understanding regional traffic and environmental conditions. Importantly, by leveraging attention rather than static graph convolutions, our model can dynamically adjust spatial influence based on context, enabling more flexible and accurate spatial reasoning.

\subsection{Spatio-Temporal Sparse Global Attention}
\label{sec:sparse_global}

While local GAT layers capture immediate spatial dependencies among neighboring H3 cells, many real-world accident patterns—such as city-wide congestion or weather-induced disruptions—require modeling of long-range, non-local interactions. To achieve this without incurring the quadratic computational cost of full attention over all \(N\) nodes, we introduce a BigBird-style \textit{sparse attention} mechanism designed to efficiently integrate both local and global signals.

Given the output node embeddings \(Z \in \mathbb{R}^{N \times d}\) from the Local GAT layer, we apply a single sparse transformer block that augments each node’s representation by allowing selective long-range communication while maintaining scalability. The core idea is to impose a sparsity pattern on the attention mask such that each node can (1) attend to its spatial neighbors (as in GAT), and (2) attend globally to a small set of shared global tokens that encode city-level context.

Formally, the sparse transformer computes:
\[
  \mathrm{STA}(Z)
  = \mathrm{softmax}\left(\frac{Z W_Q\,(Z W_K)^\top}{\sqrt{d_k}} + M\right)\,(Z W_V),
\]
where \(W_Q, W_K, W_V \in \mathbb{R}^{d \times d_k}\) are learned linear projections for the query, key, and value transformations, respectively. The term \(M \in \mathbb{R}^{N \times N}\) is a binary attention mask that enforces the sparsity pattern. Specifically:
\begin{itemize}
    \item Each node is allowed to attend to its \(k=6\) immediate H3 neighbors—preserving local continuity.
    \item A fixed set of \(G\) learnable global tokens are inserted into \(Z\), and these tokens are fully connected to all nodes. This facilitates long-range communication and global context aggregation.
\end{itemize}

This attention design ensures that information can propagate across the entire graph via a few lightweight hops—local attention routes information within a neighborhood, while global tokens act as shared intermediaries that gather and redistribute city-wide patterns. The use of sparse attention also significantly reduces memory and compute overhead compared to dense attention, especially as the number of nodes \(N\) scales.

The output of this block is a set of refined node embeddings \(\{z_i'\}_{i=1}^N\), each incorporating both their local spatial structure and long-range dependencies. We also apply residual connections and LayerNorm after the attention step to stabilize training and enhance gradient flow, following transformer best practices.

This hybrid local-global attention framework is crucial for capturing the hierarchical structure of urban environments—where accidents may be influenced by both proximal factors (e.g., nearby traffic) and distal causes (e.g., a city-wide road closure or extreme weather event occurring miles away).

\subsection{Uncertainty-Aware Risk Calibration}
\label{sec:methodology-4_5}

Accurate accident prediction systems must not only provide reliable risk scores but also quantify the confidence associated with each prediction—particularly in high-stakes, safety-critical scenarios where false positives or false negatives carry asymmetric costs. To this end, we incorporate an uncertainty-aware calibration module as the final stage of our model pipeline.

\vspace{1mm}
\paragraph{Loss Function.}
We train the final classification head—a 2-layer multilayer perceptron (MLP)—using a class-weighted binary cross-entropy loss:
\[
  \mathcal{L} = -w_1\,y\log\hat y - w_0\,(1-y)\log(1-\hat y),
\]
where \(y \in \{0,1\}\) is the ground truth label indicating the presence of an accident, \(\hat{y}\) is the predicted probability, and \(w_1 > w_0\) are weighting coefficients inversely proportional to class frequencies. Since accidents are rare events (as illustrated in Figure~\ref{fig:class_distribution}), this weighting scheme emphasizes correct identification of risky events and helps combat class imbalance by increasing recall.

\vspace{1mm}
\paragraph{Monte Carlo Dropout for Epistemic Uncertainty.}
To estimate the model's confidence in its predictions, we employ Monte Carlo (MC) Dropout—a widely adopted Bayesian approximation technique that quantifies epistemic uncertainty. Specifically, we keep dropout layers active during inference (with dropout rate \(p = 0.2\)) and perform \(K = 10\) stochastic forward passes for each input. Each pass yields a different prediction \(y_i\), and we compute the final prediction and uncertainty as:
\[
    \hat y = \frac{1}{K} \sum_{i=1}^K y_i,
    \quad
    \sigma = \sqrt{\frac{1}{K} \sum_{i=1}^K (y_i - \hat y)^2},
\]
where \(\hat{y}\) is the mean prediction representing the final risk score, and \(\sigma\) is the standard deviation capturing uncertainty. We additionally report a 95\% confidence interval for interpretability:
\[
    \hat{y} \pm 1.96\,\sigma.
\]

This approach allows us to flag uncertain predictions—those with wide confidence intervals—as lower confidence, guiding downstream decision-making (e.g., human-in-the-loop review or conservative intervention). Compared to full Bayesian neural networks, MC Dropout is lightweight and easy to integrate, requiring no major changes to the model architecture or training pipeline.

\vspace{1mm}
\paragraph{Impact.}
Incorporating this calibration step enables our system to go beyond deterministic outputs and provide interpretable confidence bounds, which is critical in real-world deployments such as emergency dispatch or infrastructure planning. As we show in Section~\ref{subsec:ablation}, this mechanism significantly improves calibration (as measured by ECE) without sacrificing accuracy, making our predictions not only strong but trustworthy.

\section{Experiments and Results}
\label{sec:experiment}

Building on our adaptive long-context framework, we conduct a series of experiments aimed at evaluating its efficacy from multiple perspectives. Specifically, we investigate: (1) incremental improvements from each architectural component via progressive ablation, (2) comprehensive benchmarking against a wide array of state-of-the-art baselines, and (3) generalization performance when transferring the model to previously unseen cities with minimal fine-tuning. This multifaceted evaluation highlights the advantages of our design choices in real-world deployment scenarios.

\subsection{Experimental Setup}

All variants are trained using 6× NVIDIA H100 GPUs (80 GB memory each), enabling large-scale multimodal training. We employ a global batch size of 12,288 (nodes × time windows), reflecting the full spatial-temporal context across cities. The adaptive window length \(w \in \{1\,\mathrm{h},\,3\,\mathrm{h},\,6\,\mathrm{h}\}\) is selected dynamically via our gating mechanism, allowing the model to optimize historical context based on local conditions. Training is conducted for 40 epochs using the AdamW optimizer with separate learning rates for each modality: \(\text{LR}_{\mathrm{ContiFormer}}=7.5\times10^{-4}\), \(\text{LR}_{\mathrm{ViT}}=1.5\times10^{-5}\), and weight decay set to \(1\times10^{-4}\). 

To comprehensively evaluate performance, we report \textbf{Accuracy}, \textbf{F1 Score}, \textbf{Precision}, \textbf{Recall}, and \textbf{Expected Calibration Error (ECE)}. ECE is especially important in safety-critical domains like accident prediction, where model confidence must align with real-world risks.

\subsection{Progressive Ablation}
\label{subsec:ablation}

To evaluate the contribution of each architectural component in our adaptive long-context framework, we conduct a stepwise ablation study. Starting from a minimal baseline, we progressively introduce the core modules described in Section~\ref{sec:methodology}, observing their effect on both classification performance (F1) and confidence calibration (ECE).

\textbf{Baseline Model.} Our starting point consists of a fixed-context architecture that combines the numerical encoder (ContiFormer) and the visual encoder (T2T-ViT), processing a static 3-hour window. The two modality streams are processed independently with no spatial graph, fusion, or adaptivity.

\vspace{2mm}
\noindent We then incrementally add the following modules:

\begin{itemize}
    \item \textbf{Local GAT:} Incorporates spatial structure by constructing an H3-based spatio-temporal graph and applying a Graph Attention Network (GAT) over fused embeddings of neighboring regions. This enhances local spatial awareness and improves inter-region accident risk propagation.
    
    \item \textbf{Cross-Modal Fusion:} Enables interaction between numerical and visual tokens via bidirectional cross-attention layers. This allows the model to reason jointly over temporal signals and spatial context, producing richer node-level embeddings.
    
    \item \textbf{Sparse Global Attention:} Extends the receptive field by allowing each node to attend globally via a BigBird-style sparse attention mechanism, while retaining neighborhood locality. This captures broader urban interactions and long-range dependencies.
    
    \item \textbf{MC-Dropout Calibration:} Introduces uncertainty modeling through Monte Carlo dropout at inference time, yielding better-calibrated probability estimates for high-risk events.
    
    \item \textbf{Adaptive Gating:} Dynamically adjusts the length of the temporal window \(w \in \{1, 3, 6\}\) for each node based on a learned volatility score computed from both numerical and visual token streams. This allows the model to use more historical context in volatile regions while conserving compute in stable ones.
\end{itemize}

\vspace{2mm}
Table~\ref{tab:ablation} summarizes the results. Each successive component leads to measurable improvements in F1, with consistent reductions in ECE. Notably, adaptive gating yields the largest performance gain, confirming the importance of tailoring temporal context to local volatility. The final model, incorporating all modules, achieves the best performance—validating the effectiveness of our modular design.

\begin{table}[h]
\centering
\caption{Ablation results: F1 / ECE ($\downarrow$)}
\label{tab:ablation}
\begin{tabular}{l|c|c}
\toprule
\textbf{Variant}                      & \textbf{F1} & \textbf{ECE} \\
\midrule
Baseline                              & 0.82        & 0.12         \\
+ Local GAT                           & 0.84        & 0.11         \\
+ Cross-Modal Fusion                  & 0.86        & 0.10         \\
+ Sparse Global Attention             & 0.88        & 0.08         \\
+ MC-Dropout Calibration              & 0.88        & 0.05         \\
+ Adaptive Gating (complete pipeline)  & \textbf{0.92} & \textbf{0.04} \\
\bottomrule
\end{tabular}
\end{table}

\subsection{Comparison to State-of-the-Art}
\label{subsec:sota}

We compare our full adaptive long-context model (\textbf{ALCo-FM}) against a comprehensive set of state-of-the-art baselines drawn from the spatio-temporal forecasting and traffic accident prediction literature. These include classical GNN-based architectures (e.g., DCRNN, STGCN), attention-based temporal models (e.g., ASTGCN, SST-GCN), and recent neural architecture search or multimodal variants (e.g., AutoSTG, VSTGCN, GLST-TARP). For a fair comparison, all baseline models operate over a fixed 3-hour temporal window and do not leverage visual context or adaptive fusion.

Table~\ref{tab:sota} presents the performance of these baselines on the complete dataset spanning 15 U.S. cities. We report \textbf{Accuracy}, \textbf{F1 Score}, \textbf{Precision}, and \textbf{Recall}. Our model achieves the highest F1 score of 0.92, outperforming all other methods by a margin of +1–2 percentage points (pp). Additionally, it achieves high Precision (0.91) and Recall (0.93), reflecting a balanced capability to detect rare accident events while minimizing false alarms.

Although \textbf{SST-GCN} reports the highest \textit{Accuracy} (0.95), it is crucial to note that this metric can be misleading in the presence of severe class imbalance, which is inherent in our dataset (i.e., the vast majority of time windows do not contain accidents). A model that overpredicts the dominant “no accident” class may achieve high accuracy without actually detecting meaningful incidents. In contrast, the \textbf{F1 score} captures the harmonic mean of Precision and Recall, making it a more suitable measure of effectiveness under imbalance, especially when identifying rare but critical safety events is the primary goal.

Our model’s superior F1 score reflects its ability to correctly identify positive accident instances while maintaining both high sensitivity (Recall) and specificity (Precision). Notably, while some baselines like GLST-TARP and FC-STGNN perform well in terms of accuracy and precision, their recall rates are lower than ours, indicating missed accident cases. In real-world deployment scenarios—such as city planning, autonomous driving, or emergency response—such false negatives could have serious consequences.

In summary, ALCo-FM not only surpasses prior models in F1 performance but also offers a reliable trade-off between detection sensitivity and false positive rate. These results demonstrate the advantage of incorporating adaptive context, multimodal fusion, and spatial reasoning in forecasting rare urban safety events.

\begin{table}[h]
\centering
\setlength{\tabcolsep}{4pt}
\caption{Performance on 15 Cities}
\label{tab:sota}
\begin{tabular}{l|cccc}
\toprule
\textbf{Method} & \textbf{Accuracy} & \textbf{F1} & \textbf{Precision} & \textbf{Recall} \\
\midrule
DCRNN                      & 0.55 & 0.46 & 0.42 & 0.50 \\
STGCN                      & 0.64 & 0.49 & 0.51 & 0.47 \\
ASTGCN                     & 0.74 & 0.60 & 0.58 & 0.63 \\
GWN                        & 0.72 & 0.53 & 0.55 & 0.52 \\
STSGCN                     & 0.85 & 0.72 & 0.78 & 0.67 \\
STFGNN                     & 0.82 & 0.65 & 0.66 & 0.64 \\
STGODE                     & 0.84 & 0.65 & 0.59 & 0.72 \\
AutoSTG                    & 0.71 & 0.49 & 0.50 & 0.48 \\
Auto-DSTSG                 & 0.73 & 0.59 & 0.63 & 0.55 \\
SST-DHL                    & 0.89 & 0.78 & 0.85 & 0.72 \\
VSTGCN                     & 0.80 & 0.74 & 0.78 & 0.70 \\
SST-GCN                    & \textbf{0.95} & 0.80 & 0.89 & 0.73 \\
DGCRN                      & 0.89 & 0.72 & 0.74 & 0.71 \\
HetConvLSTM                & 0.66 & 0.48 & 0.52 & 0.45 \\
DSTGCN                     & 0.84 & 0.66 & 0.67 & 0.65 \\
MVMT-STN                   & 0.79 & 0.59 & 0.60 & 0.59 \\
UTAASTRL                   & 0.81 & 0.59 & 0.58 & 0.61 \\
UTARPR                     & 0.90 & 0.76 & 0.66 & 0.89 \\
GLST-TARP                  & 0.92 & 0.85 & 0.83 & 0.88 \\
FC-STGNN                   & 0.94 & 0.79 & 0.81 & 0.78 \\
\midrule
\textbf{ALCo-FM}           & 0.94 & \textbf{0.92} & \textbf{0.91} & \textbf{0.93} \\
\bottomrule
\end{tabular}
\end{table}

\subsection{Generalization to Unseen Cities}
\label{subsec:unseen}

To evaluate the robustness and transferability of our model, we assess its performance on three previously unseen cities: Columbus, Portland, and Oklahoma City. These cities were completely excluded from the training set to simulate realistic deployment in new geographic regions. After pretraining our model on the remaining 15 cities, we fine-tune only the final GAT layer and the MLP classification head for each held-out city, keeping all other layers frozen. This lightweight fine-tuning is conducted for just 5 epochs with early stopping based on validation F1.

Table~\ref{tab:unseen} reports the model’s performance across standard metrics. Remarkably, despite the limited number of trainable parameters and short fine-tuning duration, our framework consistently achieves strong F1 scores across all three cities. High precision and recall further indicate that the model adapts effectively to novel spatial distributions and demographic profiles. 

This outcome demonstrates the power of our shared spatio-temporal representations, which generalize well across diverse urban regions. In particular, the model retains the ability to correctly identify rare accident events in unfamiliar cities, while keeping the false positive rate low—an essential feature for safety-critical systems.

Moreover, the fine-tuning process reflects a practical use case in which a pre-trained model is deployed to a new city with limited additional training. Since we update only a small portion of the model, this process is both compute-efficient and data-efficient, making it suitable for real-time adaptation scenarios, edge deployment, or transfer learning in resource-constrained municipalities.

\begin{table}[h]
\centering
\caption{Fine-Tuning on Unseen Cities}
\label{tab:unseen}
\begin{tabular}{l|cccc}
\toprule
City            & Accuracy & F1   & Precision & Recall \\
\midrule
Columbus        & 0.92     & 0.90 & 0.89      & 0.91   \\
Portland        & 0.93     & 0.90 & 0.88      & 0.93   \\
Oklahoma City   & 0.90     & 0.89 & 0.88      & 0.91   \\
\bottomrule
\end{tabular}
\end{table}

Overall, these results highlight our model’s strong generalization capabilities, reinforcing its value as a plug-and-play foundation for large-scale traffic accident prediction across heterogeneous urban environments.

\section{Conclusion and Future Work}
\label{sec:conclusion}

We present ALCo-FM, an adaptive long-context foundation model that dynamically selects history lengths via uncertainty-driven H3-hexagon aggregation and fuses numerical, environmental, and spatial-temporal signals within a single backbone. Evaluated on 1,771 regions across 15 U.S. cities, ALCo-FM achieves 0.94 accuracy and 0.92 F1—surpassing over 20 strong baselines—and generalizes to three unseen cities with minimal fine-tuning. Its uncertainty-aware design delivers not only state-of-the-art performance but also well-calibrated risk estimates, making it both interpretable and reliable for real-world deployments. Future work will explore adaptive spatial indexing beyond fixed H3 grids, incorporate telematics and mobile-sensor streams, and develop cross-city domain-adaptation techniques to further bolster robustness and generalization.


\bibliography{example_paper}
\bibliographystyle{icml2025}

\newpage
\appendix

\clearpage

\onecolumn
\setcounter{page}{1}

\section*{Appendix}
\label{sec:appen}

\section{Details of Dataset Descriptions}
\label{apn:appen_C}

\subsection{Traffic Events Data}
\label{app:App_Btraffic}

To build a robust dataset for accident risk prediction, we leverage a comprehensive car accident dataset by Moosavi et al.~\cite{moosavi2019countrywide}, covering 49 states across the United States. This dataset, collected from February 2016 to March 2023, comprises approximately 7.7 million accident records. The data is aggregated from multiple sources, including Bing, MapQuest, real-time traffic APIs that collect information from transportation departments, law enforcement agencies, traffic cameras, and in-road traffic sensors.

The dataset provides a rich set of attributes for each recorded accident, such as precise time and location of occurrence (including city, state, and ZIP code), severity levels, duration, and the length of resulting traffic impact. These features enable a granular analysis of traffic incidents, offering insights into possible patterns and contributing factors. The accident data is continuously updated in real time through multiple Traffic APIs, providing high temporal resolution and nationwide coverage of the contiguous United States.
In addition to core variables like time, location, and severity, each record also includes relevant ``Point of Interest (POI)'' annotation tags, sourced from OpenStreetMap (OSM)~\cite{OpenStreetMap}. Table~\ref{tab:poi_description} presents an overview of the different POI categories found in the dataset, along with their corresponding descriptions.

\begin{table*}[h]
    \centering
    \caption{Definition of Point-Of-Interest (POI) annotation tags based on OpenStreetMap (OSM).}
    \label{tab:poi_description}
  \setlength{\extrarowheight}{2pt}
  \renewcommand{\arraystretch}{1.2}
    \begin{tabular}{@{} l | p{0.82\linewidth} @{}} 
        \toprule
        \textbf{Type} & \textbf{Description} \\
        \midrule
        Amenity & Denotes specific locations such as restaurants, libraries, colleges, bars, etc. \\
        Bump & Represents speed bumps or humps designed to reduce vehicle speed. \\
        Crossing & Indicates designated pedestrian or cyclist crossings across roads. \\
        Give-way & Road sign that dictates priority at intersections. \\
        Junction & Represents highway ramps, exits, or entry points. \\
        No-exit & Marks a point where travel cannot continue further along a designated path. \\
        Railway & Identifies locations where railway tracks are present. \\
        Roundabout & Indicates a circular road junction facilitating smooth traffic flow. \\
        Station & Denotes public transportation hubs such as bus stops or metro stations. \\
        Stop & Signifies stop signs at intersections or along roads. \\
        Traffic Calming & Encompasses road features designed to slow down vehicle speed. \\
        Traffic Signal & Represents traffic lights at intersections or pedestrian crossings. \\
        Turning Loop & Designates widened sections of a highway featuring a non-traversable island for turning. \\
        \bottomrule
    \end{tabular}
\end{table*}


\subsection{Demographic Attributes Data}
\label{app:App_Bdemo}
To account for socioeconomic and population-based factors, we incorporate a comprehensive demographic dataset sourced from an online geographic data resource providing detailed postal and demographic information across the U.S. \cite{us_zipcodes}.  It aggregates authoritative data from reputable sources such as the U.S. Census Bureau~\cite{us_census} and the United States Postal Service (USPS)~\cite{usps_zipcodes}, ensuring the accuracy and reliability of the information presented. The platform offers extensive datasets, including ZIP code boundaries, population statistics, income levels, housing data, and geographic coordinates, making it a valuable tool for researchers, policymakers, and businesses.
The dataset encompasses approximately 45,000 U.S. ZIP Codes and contains 150 demographic variables, offering a multi-faceted perspective on local socio-economic conditions. The data collection process involved API-based retrieval methods, ensuring a wide coverage of geographical regions while maintaining data consistency and accuracy.

Each ZIP code entry in the dataset includes information across four primary categories: (1) Population Characteristics, covering attributes such as age distribution, gender ratios, and racial composition; (2) Housing Information, detailing aspects such as homeownership rates, rental statistics, and average household sizes; (3) Employment and Income Statistics, providing insights into median household income, employment rates, and occupational distribution; and (4) Education Levels, offering data on educational attainment and literacy rates within each region.

By leveraging this rich demographic information, our study aims to analyze the potential influence of socio-economic factors on traffic patterns, accident frequencies, and severity levels. The inclusion of such granular data enhances our model's ability to capture regional disparities and provide more accurate predictive insights.


\subsection{Weather Conditions Data}
\label{app:App_Bweather}

Since meteorological factors significantly influence traffic flow and accident likelihood, incorporating weather data is crucial for developing robust accident prediction models. Weather conditions such as temperature, precipitation, wind speed, and visibility can directly impact driving behavior, road surface conditions, and vehicle performance. Adverse weather, including rain, snow, fog, and extreme temperatures, can increase the risk of accidents by reducing visibility, affecting braking distances, and leading to hazardous road conditions. Moreover, sudden weather changes can disrupt normal traffic patterns, leading to congestion and increased accident probabilities. Hence, integrating meteorological data allows for more accurate and context-aware accident prediction models.

For our study, we obtained hourly weather data spanning from 2016 to 2023 for selected stations from the Iowa State University Iowa Environmental Mesonet (IEM) - ASOS Network ASOS-AWOS-METAR Data Download platform~\cite{iem_asos}. The IEM archives automated airport weather observations from various global locations, primarily collected from Automated Surface Observation System (ASOS) and Automated Weather Observation System (AWOS) stations. These observations provide detailed insights into prevailing atmospheric conditions at a high temporal resolution. The processed weather dataset was structured as a time series spanning the study period, with each row representing an hourly observation for the respective stations. This structured representation enables downstream modeling tasks to analyze temporal correlations between weather patterns and accident occurrences effectively.

By leveraging this comprehensive weather dataset, our accident prediction framework integrates meteorological variables alongside traffic and demographic data to capture the complex interplay between environmental conditions and accident risks. The inclusion of weather data enriches model accuracy by accounting for seasonal variations, adverse weather events, and localized microclimates that might otherwise be overlooked in traditional predictive models. Recent research has also explored the utility of deep learning for short-term weather-dependent prediction tasks—such as solar radiation forecasting in the Pacific Northwest—demonstrating how weather variability can directly influence decision-making in real-time systems~\cite{neogi2023deep}. Table~\ref{tab:weather_features} contains the list of weather features we collected.

\begin{table*}[ht!]
  \centering
  \caption{Description of Weather Features}
  \label{tab:weather_features}
  \setlength{\extrarowheight}{2pt}
  \renewcommand{\arraystretch}{1.2}
  \begin{tabular}{@{} l | p{12cm} @{}}
    \toprule
    \textbf{Feature} & \textbf{Description} \\
    \midrule
    \textbf{tmpf}   & Air Temperature in Fahrenheit, typically measured at 2 m above the ground. \\
    \textbf{dwpf}   & Dew Point Temperature in Fahrenheit, typically measured at 2 m above the ground. \\
    \textbf{relh}   & Relative Humidity expressed as a percentage. \\
    \textbf{drct}   & Wind Direction in degrees, measured from true north. \\
    \textbf{sknt}   & Wind Speed in knots. \\
    \textbf{p01i}   & One-hour precipitation amount in inches, recorded from the observation time to the previous hourly precipitation reset. May include melted frozen precipitation. \\
    \textbf{alti}   & Pressure altimeter measurement in inches. \\
    \textbf{mslp}   & Sea Level Pressure measured in millibars. \\
    \textbf{vsby}   & Visibility in miles. \\
    \textbf{skyc1}  & Sky Level 1 Coverage. \\
    \bottomrule
  \end{tabular}
\end{table*}


\subsection{Map Image Representation}
\label{app:App_Bimg}
To capture the spatial context and road-network features associated with accident locations, we utilize a systematic approach based on hexagonal spatial segmentation and map image retrieval. Each accident location is mapped to a unique hexagonal region using Uber’s Hexagonal Hierarchical Spatial Indexing (H3) ~\cite{uber_h3} with a resolution of \( R = 7 \). This resolution results in hexagons with an approximate edge length of 2,604 meters and a total area of about 5.16 km\(^2\), effectively covering the vicinity of the accident site.

Once the hexagonal zoning is established, the center coordinates of each hexagonal region are used to retrieve corresponding map tiles from OpenStreetMap (OSM) ~\cite{OpenStreetMap}. Specifically, we extract square map tiles at a zoom level of 14, which results in images of size \( 256 \times 256 \) pixels. At this zoom level, each pixel represents approximately 9.547 meters, making the total coverage of each tile approximately 2.44 km per side, corresponding to an area of 5.95 km\(^2\). The spatial coverage of these map tiles ensures that the entire hexagonal zone is sufficiently represented, allowing the model to incorporate comprehensive geographic features.

\begin{figure*}[h!]
    \centering
    \begin{minipage}{\textwidth}
        \centering
        \includegraphics[height=3.15cm]{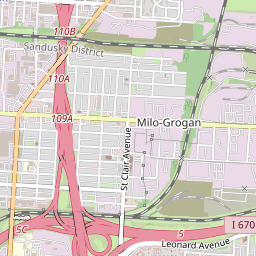}
        \hspace{2mm}
        \includegraphics[height=3.15cm]{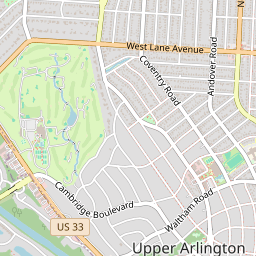}
        \hspace{2mm}
        \includegraphics[height=3.15cm]{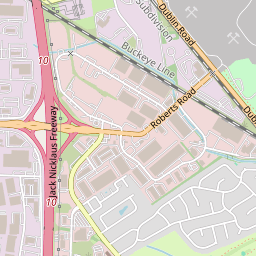}
        \hspace{2mm}
        \includegraphics[height=3.15cm]{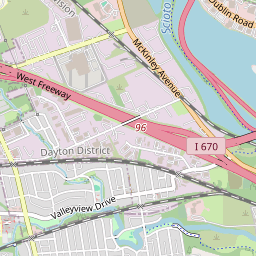}
    \end{minipage}
    \caption{Examples of map tiles obtained from OSM (zoom level = 14), representing approximate H3 zones (R = 7) in Columbus, Ohio}
    \label{fig:four_images}
\end{figure*}

Figure~\ref{fig:four_images} presents some examples of map tiles retrieved from OSM at zoom level 14, which are used to approximate H3 zones (R = 7) in regions of Columbus, Ohio. These map images encapsulate critical environmental elements, including roads, intersections, buildings, and other infrastructure, offering valuable insights into potential accident-prone areas~\cite{10.1145/3557915.3560943, 10.1145/3615900.3628769}. By leveraging these geospatial insights with rich textural data (e.g., road names), our model can identify spatial patterns, such as the density of road networks, the complexity of intersections, and the presence of high-traffic zones, which contribute to the occurrence of traffic incidents. This multi-faceted spatial encoding provides an essential context for enhancing the accuracy of our accident prediction framework. Similar to how multimodal models in Document VQA have begun to integrate spatial localization for enhanced interpretability~\cite{mohammadshirazi2024dlavadocumentlanguagevision, mohammadshirazi2024docparsenet}, our spatial reasoning leverages geographic features to ground model predictions more transparently and contextually.

\section{Dataset Characterization and Statistical Analysis}
\label{apn:appen_dataset_analysis}

To address any concerns regarding the perceived simplicity of the dataset, we provide a comprehensive statistical and structural analysis. The following figures highlight the underlying heterogeneity, feature complexity, and imbalanced nature of the accident prediction problem—underscoring that the dataset reflects real-world challenges rather than a synthetic or overly simplified scenario.

\subsection{Feature Correlation and Redundancy}
\begin{figure}[h]
    \centering
    \includegraphics[width=1\linewidth]{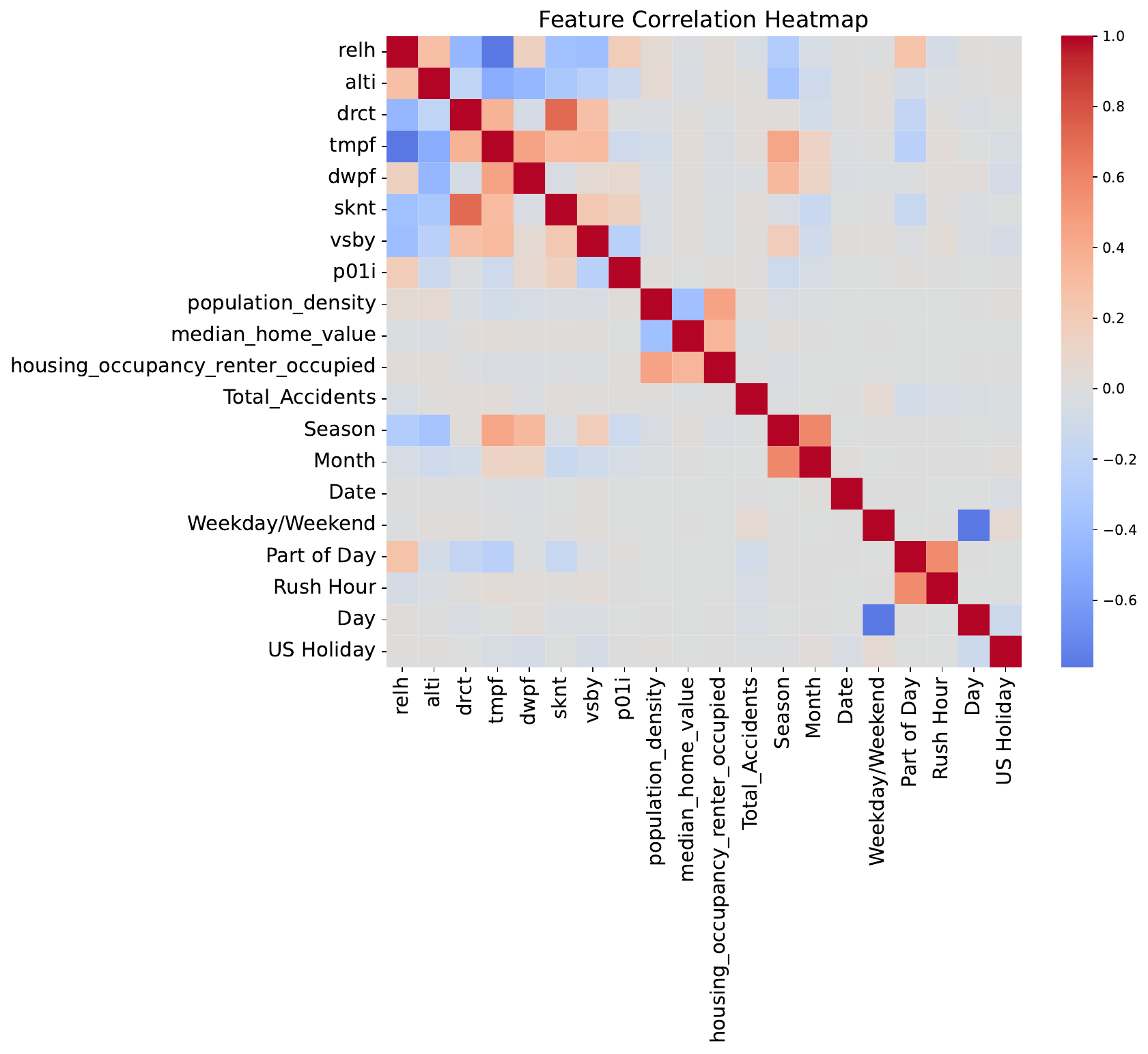}
    \caption{Pairwise correlation matrix of numerical and categorical features.}
    \label{fig:corr_heatmap}
\end{figure}

Figure~\ref{fig:corr_heatmap} presents a feature correlation heatmap showing the pairwise Pearson correlation coefficients among all structured features. The presence of low to moderate correlations across most features indicates that the dataset is neither redundant nor trivially separable. Notably, some weather attributes (e.g., temperature, humidity, wind speed) show weak or no linear relationships with accident occurrence, emphasizing the need for nonlinear modeling strategies like GNNs and Transformers. Temporal and demographic features also remain weakly correlated, supporting the multimodal learning requirement.

\subsection{Class Imbalance Across Cities}
\begin{figure}[h]
    \centering
    \includegraphics[width=\linewidth]{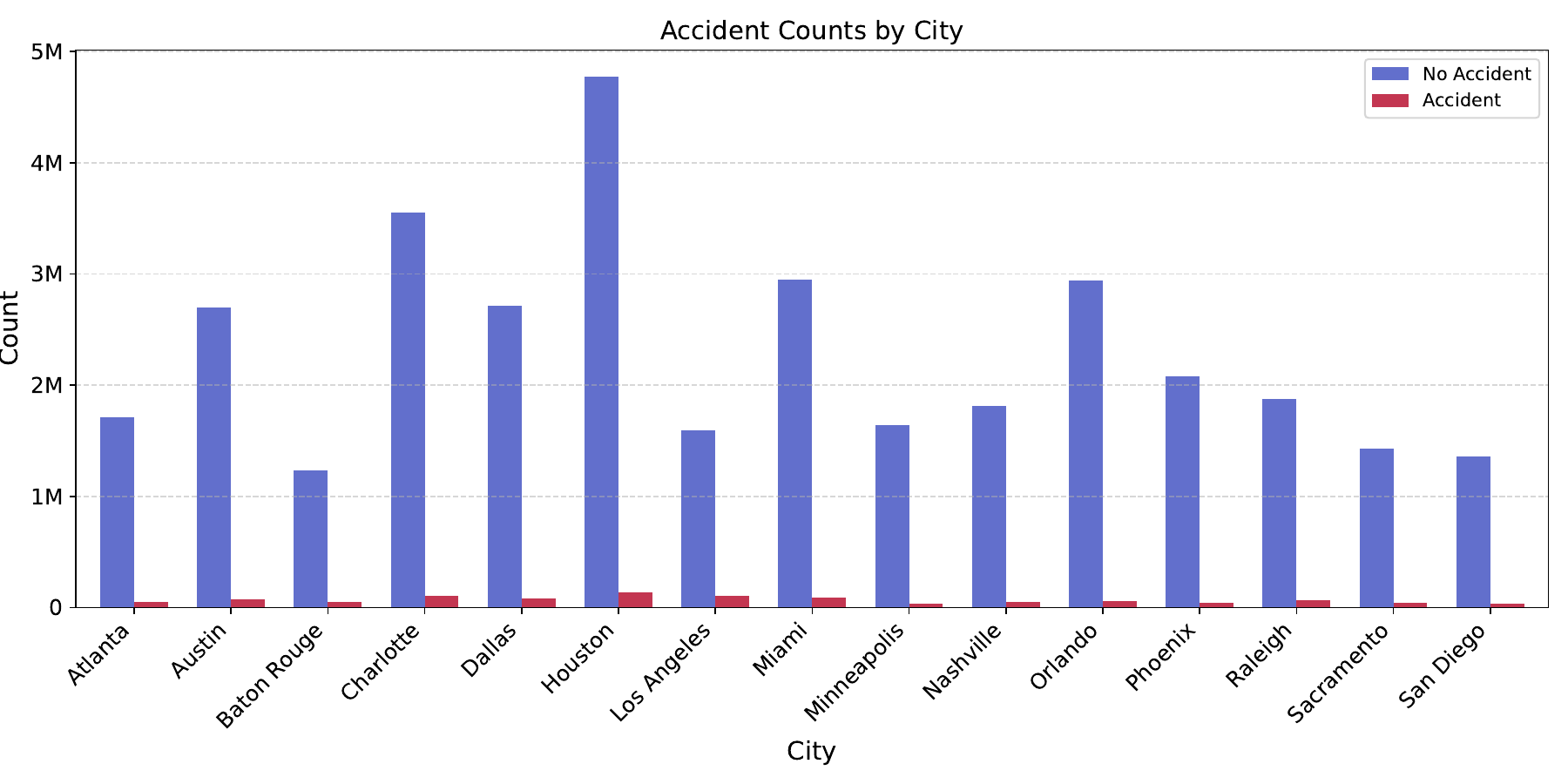}
    \caption{Distribution of accident vs. no-accident instances across different cities.}
    \label{fig:class_distribution}
\end{figure}

As illustrated in Figure~\ref{fig:class_distribution}, the dataset is highly imbalanced, with a significant dominance of “no accident” records across all cities. This mirrors real-world scenarios where accidents are rare events, despite dense traffic volumes. The imbalance varies across cities, requiring the model to generalize under skewed label distributions—a challenge not typically encountered in toy or synthetic datasets.

\subsection{Feature Complexity and Dimensionality}
\begin{figure}[h]
    \centering
    \includegraphics[width=1\linewidth]{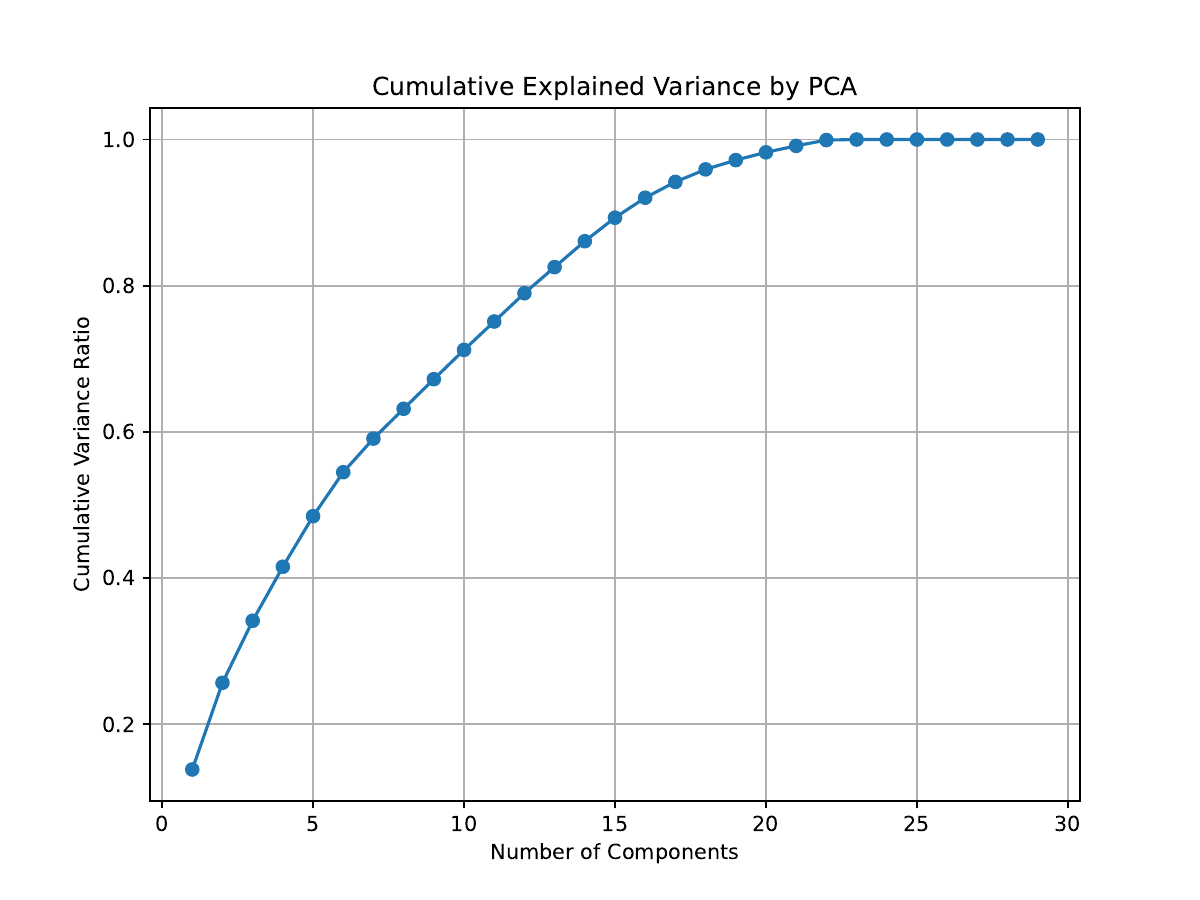}
    \caption{Cumulative explained variance from PCA across top 30 components.}
    \label{fig:pca_variance}
\end{figure}

To assess the dimensional structure of the data, we apply Principal Component Analysis (PCA). Figure~\ref{fig:pca_variance} shows that over 20 principal components are needed to capture 95\% of the variance, indicating that the dataset is high-dimensional and not compressible into a simple subspace. This further confirms the necessity of deep learning-based feature extraction as opposed to shallow models or linear classifiers.

\subsection{Feature Importance Distribution}
\begin{figure}[h]
    \centering
    \includegraphics[width=1\linewidth]{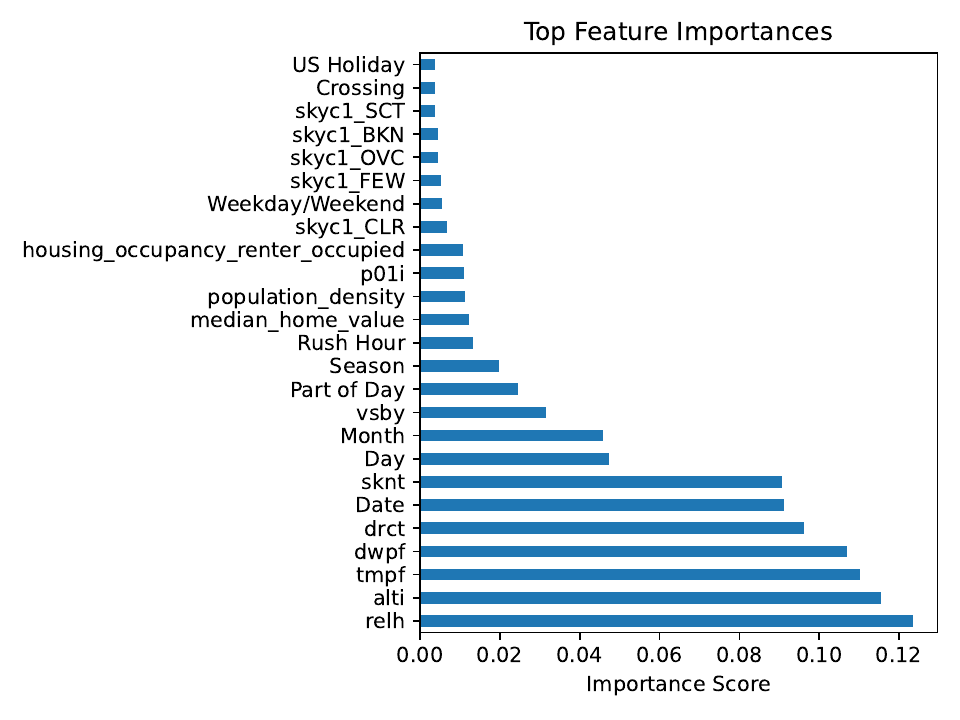}
    \caption{XGBoost-based feature importance ranking.}
    \label{fig:feature_importance}
\end{figure}

Figure~\ref{fig:feature_importance} ranks features by their importance scores using an XGBoost classifier. A wide range of features—including weather attributes (e.g., temperature, humidity), temporal segments (e.g., date, rush hour), and demographic variables—contribute meaningfully to the accident prediction task. This confirms the dataset's multimodal nature and the nontrivial interaction between features, invalidating any assumptions of simplicity.

\subsection{Nonlinear Separability in Latent Space}
\begin{figure}[h]
    \centering
    \includegraphics[width=1\linewidth]{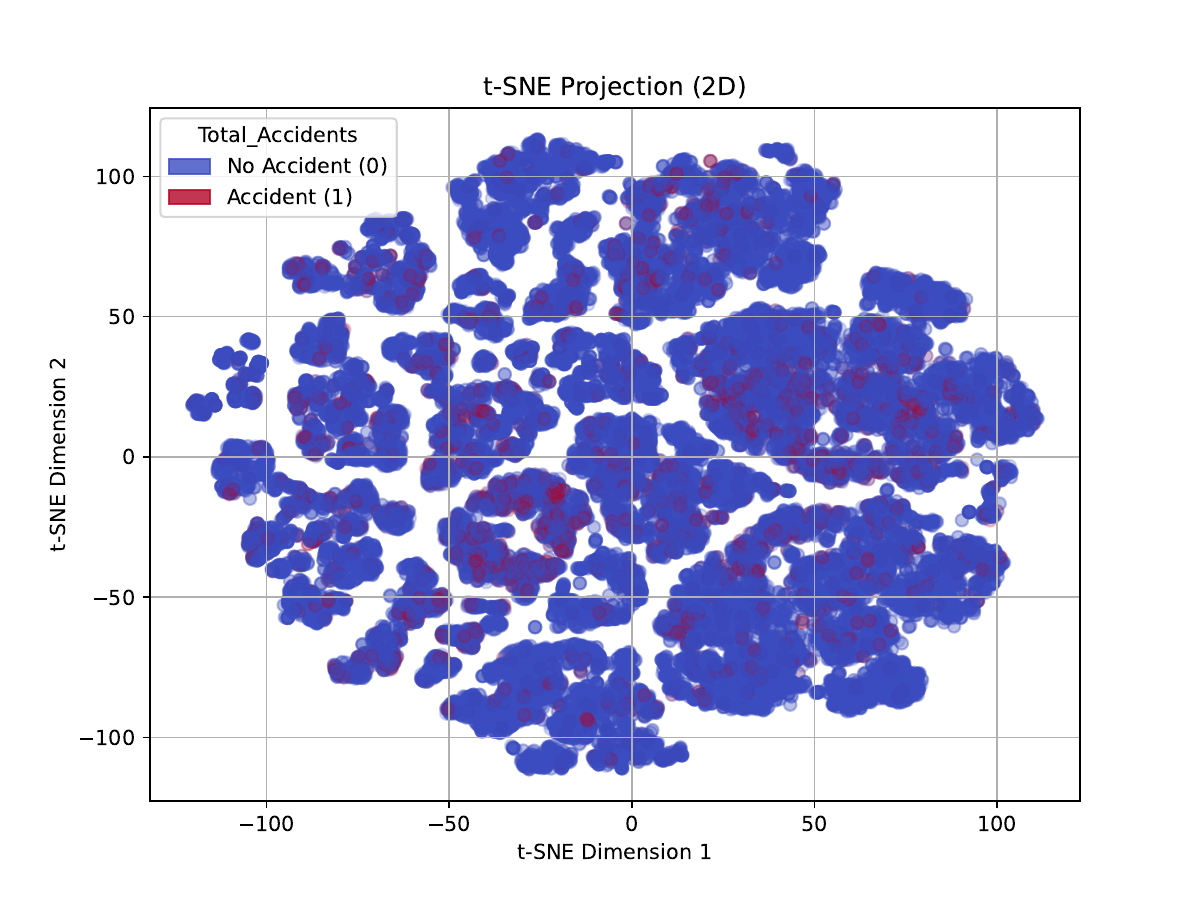}
    \caption{t-SNE projection of the dataset in 2D space, colored by accident labels.}
    \label{fig:tsne}
\end{figure}

\begin{figure}[h]
    \centering
    \includegraphics[width=1\linewidth]{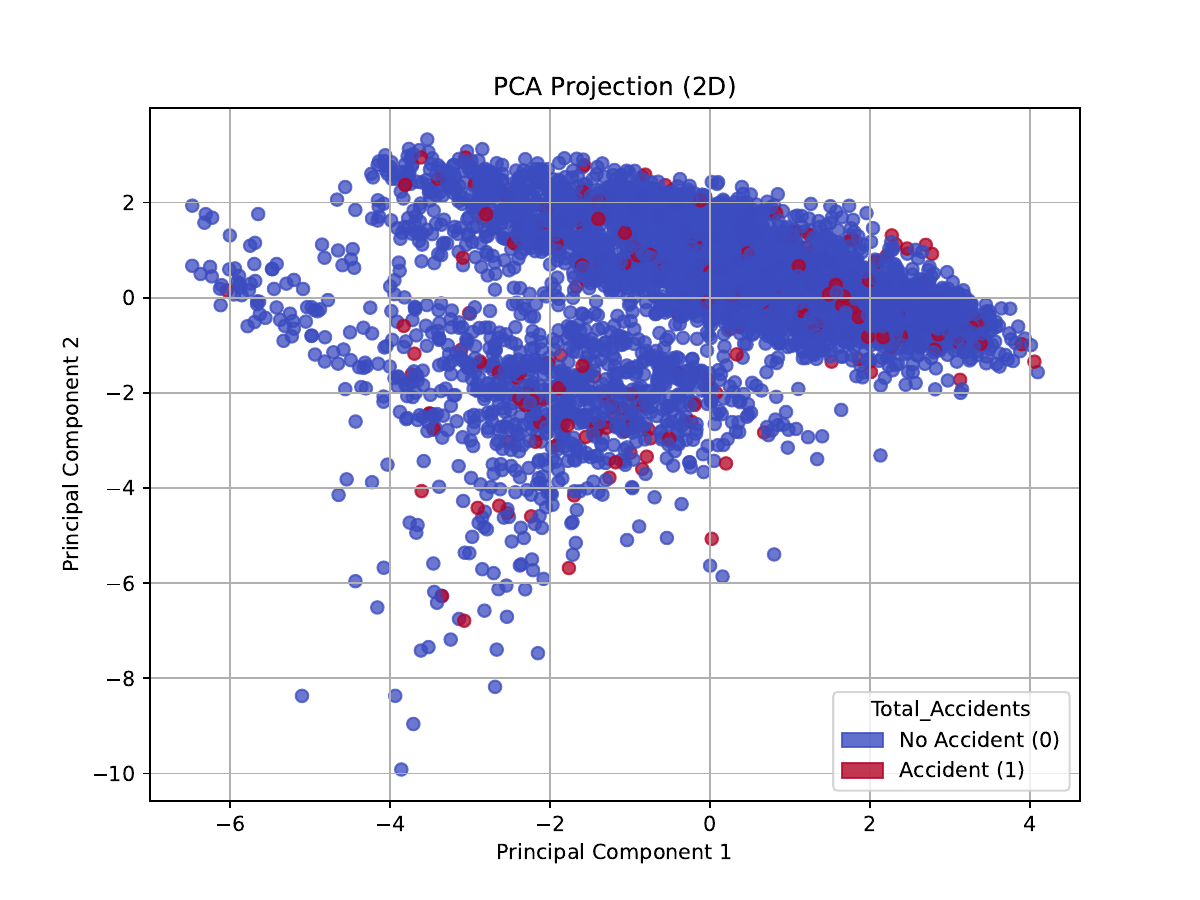}
    \caption{PCA projection of the dataset in 2D space, colored by accident labels.}
    \label{fig:pca_projection}
\end{figure}

Figures~\ref{fig:tsne} and \ref{fig:pca_projection} show 2D projections of the high-dimensional dataset using t-SNE and PCA respectively. Both visualizations reveal that accident vs. non-accident data points are not linearly separable and form dense, overlapping clusters. The sparse presence of accident cases amidst abundant “no accident” points highlights the difficulty of learning decision boundaries and underscores the need for advanced modeling techniques capable of capturing spatio-temporal dependencies and multimodal feature interactions.

Thus, the dataset presents real-world challenges such as high-dimensionality, severe class imbalance, weak feature correlations, and complex nonlinear boundaries. These characteristics necessitate the use of sophisticated architectures like ALCo-FM and invalidate concerns regarding dataset simplicity. This appendix offers transparent evidence that accident prediction in our context is a challenging and realistic task.


\section{Details of Numerical Data Preprocessing and Feature Engineering}
\label{apn:appen_A}

\subsection{Merging the Data Sources}
\label{app:App_C1}
We start by loading the Traffic Event Data, which provides detailed accident records including ZIP codes and geographic coordinates (latitude, longitude). Next, we load the Demographic Attributes Data, keyed by ZIP codes. 

Using the shared ZIP code field, we perform a join operation to enrich each accident record with socio-economic attributes (e.g., population density, income levels, and additional demographic features). This integration is crucial for capturing the broader context in which accidents occur, ensuring that each record in the traffic event dataset is accompanied by pertinent demographic indicators. The merged dataset thus combines spatio-temporal accident details with socio-economic factors, serving as a foundation for subsequent filtering and feature engineering steps.

\subsection{City Selection Based on Missing Data Ratio}
\label{app:App_C2}
Upon investigating the dataset, we found that missing values are scattered across various records and feature columns, rather than being concentrated in specific rows or attributes. This pattern suggests that there are no dominant bulk-missing features or entirely incomplete records, making it suitable to measure missingness at the city level and apply a filtering criterion accordingly.

A common strategy in data science for handling significant missingness is to measure the fraction of all possible data entries that are missing and then filter out entities (in this case, cities) with excessive missingness. Let \(R_c\) be the total number of records for city \(c\), and let \(D\) be the total number of columns after merging all relevant attributes. Define \(M_c\) as the sum of missing feature values across all records in city \(c\). Formally, the missingness ratio \(\delta_c\) is:
\begin{equation}
    \delta_c = \frac{M_c}{R_c \times D}.
\end{equation}
We then rank the cities in ascending order of \(\delta_c\) and retain the top 15 cities with the lowest missing ratios. By eliminating cities that surpass a certain threshold of missingness, we reduce the need for heavy imputation, which can introduce additional noise and uncertainty. This filtering step thus preserves higher data fidelity and provides a more reliable foundation for subsequent analyses. In the sebsequent subsections we have shown the list of selected cities used for training the model.

\subsection{Spatial Partitioning with H3}
\label{app:App_C3}
Although demographic attributes are keyed to ZIP codes, these codes vary greatly in both shape and size, making them suboptimal for fine-grained spatial analysis. To address this limitation, we employ Uber's H3 library at a resolution of \(R = 7\) to partition the study area into a grid of hexagonal cells (referred to as Area IDs). At this resolution, each cell covers approximately 5.16~km\(^2\) and has edges of about 2.6~km, yielding a uniform spatial framework that applies consistently across all cities under consideration. 

This hexagonal partitioning offers several distinct benefits:
\begin{itemize}
    \item \textbf{Uniformity:} H3 cells are designed to be nearly identical in shape and area, facilitating more consistent distance-based comparisons than ZIP codes, which were originally developed for mail delivery rather than spatial analytics.
    \item \textbf{Neighbor Identification:} By encoding each accident’s latitude and longitude into a hexagonal cell, we can readily establish adjacency relationships among cells, which is crucial for subsequent graph-based modeling.
    \item \textbf{Scalability:} The hierarchical nature of H3 allows for flexible adjustments in spatial granularity, supporting diverse analytical resolutions without altering the fundamental grid structure.
\end{itemize}

In this study, we favor H3 over ZIP-code boundaries to avoid the irregularities and non-uniform areas that arise from using postal regions. The standardized hexagonal cells not only simplify distance calculations but also offer a consistent basis for comparing spatial relationships across different cities and states.

\subsection{Filtering by H3 Regions}
\label{app:App_C4}

Even within a single city, certain H3 cells can contain very few valid (non-missing) accident records. Excessive reliance on imputation for these sparse cells may introduce noise and bias, ultimately obscuring meaningful patterns. To mitigate this risk, we apply two additional filters:

\paragraph{(1) Minimum Record Threshold.}
We first enforce a lower bound on the number of records per H3 cell. Specifically, any cell with 
\[
    \text{TotalRecords} < 100
\]
is excluded from further analysis. This criterion ensures that each cell considered has sufficient coverage for a robust evaluation of local traffic and environmental attributes.

\paragraph{(2) Non-Missing Ratio Threshold.}
Next, we measure the proportion of records in each H3 cell that are fully complete (i.e., contain no missing values in the core fields). Formally, we define:
\begin{equation}
    \text{NonMissingRatio}_{\text{H3}} 
    \;=\; \frac{\text{NonMissingRecords}}{\text{TotalRecords}}.
\end{equation}
To maintain high data fidelity, we discard any cell with 
\[
    \text{NonMissingRatio}_{\text{H3}} < 0.95.
\]
Requiring at least 95\% of records to be fully observed limits the extent of imputation, thereby preserving the integrity of the underlying signals.

\paragraph{Outcome.}
By applying these two filters, we arrive at a final subset of H3 cells (drawn from the 15 selected cities) that is both dense (in terms of record count) and predominantly complete (in terms of observed values). This ensures that subsequent analyses focus on geographically localized regions where the data quality is sufficiently high to support reliable pattern discovery and prediction. Table \ref{tab:final_nodes} shows the final number of nodes in each selected cities.


\subsection{Time-Series Construction \& Label Definition}
\label{app:App_C5}
In this step, we transform the accident data into discrete time slices spanning \(\text{June 1, 2016}\) to \(\text{March 31, 2023}\), grouped in 3-hour intervals. The choice of a 3-hour window is guided by two principal considerations:

\begin{itemize}
    \item \textbf{Balancing sparsity and resolution:} Using a shorter interval (e.g., 1~hour) risks producing excessively sparse data in many H3 cells, as accidents do not occur frequently enough in every region. Conversely, a coarser interval (e.g., 6~hours) can obscure important intra-day variations, such as rush-hour patterns.
    \item \textbf{Worst-case severity labeling:} If multiple accidents occur within the same 3-hour window, we assign the highest observed severity to that window. This “worst-case” labeling strategy ensures that models can account for the most critical safety scenarios, thereby providing a conservative estimate for risk-focused applications.
\end{itemize}

Once grouped by these 3-hour windows, we aggregate all accidents occurring in a given H3 cell within the same time slice. The aggregated record is then enriched with additional features (e.g., weather variables), creating a comprehensive event representation for further modeling.

\subsection{Temporal Representation}
\label{app:App_C6}

Temporal attributes are pivotal for modeling accident risk, weather variations, and traffic congestion. 
Once the accident records have been aggregated into 3-hour intervals, 
we further enrich each time slice with cyclical and categorical indicators that capture intra-day, weekly, and seasonal patterns. 
Table~\ref{tab:temporal_features} summarizes the principal temporal features extracted from the dataset.


\begin{table*}[t]
  \centering
  \caption{Summary of temporal features derived from timestamps spanning 2016--2023. Numeric values used for categorical encoding are indicated in parentheses.}
  \label{tab:temporal_features}
  \setlength{\extrarowheight}{2pt}
  \renewcommand{\arraystretch}{1.2}
  \begin{tabular}{@{} l | p{0.82\linewidth} @{}}
    \toprule
    \textbf{Feature} & \textbf{Description} \\
    \midrule
    Season      & Categorized based on the month: \textbf{Winter (0)} [Dec, Jan, Feb]; \textbf{Spring (1)} [Mar, Apr, May]; \textbf{Summer (2)} [Jun, Jul, Aug]; \textbf{Fall (3)} [Sep, Oct, Nov]. \\
    Month       & Month number \textbf{(1--12)}. \\
    Date        & Day of the month \textbf{(1--31)}. \\
    Day         & Day of the week, encoded as \textbf{Monday (0)} to \textbf{Sunday (6)}. \\
    Weekday     & Binary label: \textbf{Weekday (1)} if Monday–Friday; \textbf{Weekend (0)} if Saturday or Sunday. \\
    Holiday     & Binary indicator: \textbf{Holiday (1)} if the date is a recognized U.S.\ public holiday; otherwise \textbf{Non-Holiday (0)}. \\
    Part-of-Day & \textbf{Morning (0)} [6:00–11:59]; \textbf{Afternoon (1)} [12:00–17:59]; \textbf{Evening (2)} [18:00–23:59]; \textbf{Night (3)} [00:00–5:59]. \\
    Rush-Hour   & \textbf{Morning Rush (0)} [6:00–8:59]; \textbf{Evening Rush (1)} [15:00–17:59]; \textbf{Non-Rush Hours (2)} [all other times]. \\
    \bottomrule
  \end{tabular}
\end{table*}

By incorporating these temporal markers, our predictive models gain finer‐grained insights into traffic‐related phenomena, 
enabling more accurate modeling of short‐term fluctuations, diurnal patterns, and seasonal variations 
that influence accident probabilities.

\subsection{Weather Data Integration}
\label{app:App_C7}
In order to incorporate meteorological context into our accident prediction framework, we align each 3-hour accident window with the relevant weather observations collected over the same time span. Specifically, we aggregate numerical weather variables (e.g., temperature, humidity) by computing their arithmetic mean within each 3-hour period. For the categorical variable \texttt{skyc1}, indicating cloud cover, we convert its discrete labels to an integer encoding, compute the average of these encoded values across the window, and then round back to the nearest integer to determine a single representative category. The weather dataset exhibited a minimal missingness rate of only 0.01\%, making it impractical to discard records due to sparsity. Instead, we applied bidirectional interpolation \cite{lam1983spatial} to impute missing values before aggregating weather attributes over each 3-hour window.

This procedure yields a compact set of weather attributes for each (H3 cell, 3-hour time slice) tuple, enabling downstream models to account for variations in atmospheric conditions. By jointly modeling traffic, demographic, and weather information, our framework benefits from a more holistic view of the factors influencing accident risk.

\subsection{Handling Missing Data}
\label{app:App_C8}

Despite comprehensive data collection efforts, the large-scale nature of our dataset inevitably resulted in a non-trivial percentage of missing entries. 

\paragraph{Overview of Imputation Techniques.}
Rather than discarding incomplete rows, we opted to impute missing values using several state-of-the-art algorithms. This choice balances data retention against the risk of inaccurate imputation, ensuring we utilize as much of the available information as possible.

The following methods were evaluated, with hyperparameters selected via \textbf{GridSearchCV} over an appropriate range:

\begin{itemize}
    \item \textbf{FAISS kNN (k=5) \cite{johnson2019billion}}:
    Scalable similarity search, well-suited for large datasets. 
    It addresses the slow query issue of classical kNN by using efficient nearest-neighbor lookups. We chose \( k=5 \) after tuning on \( k \in \{3, 5, 7, 9, 15\} \), as it yielded the lowest error.
    \item \textbf{SAITS \cite{Du_2023}}:
    A self-attention-based time-series imputation approach that adaptively models temporal dependencies. We fine-tuned the transformer layers and attention heads in the range \( \{1, 2, 4, 8\} \) and selected optimal values based on validation performance.
    \item \textbf{XGBoost (max\_depth=6, learning\_rate=0.05) \cite{Chen_2016}}:
    A tree-ensemble method that leverages non-linear relationships to estimate missing feature values. Hyperparameters such as \texttt{max\_depth} and \texttt{learning\_rate} were selected using GridSearch over \{3, 6, 9\} and \{0.01, 0.05, 0.1\}, respectively.
    \item \textbf{Iterative Imputation (max\_iter=10, estimator=XGBoost) \cite{Guo_2024}}:
    A model-based iterative scheme that treats each feature with missing values as a regression target,
    iteratively refining estimates using other features as predictors. We experimented with linear regression, random forests, and XGBoost as the estimator, finding XGBoost to perform best.
\end{itemize}

\begin{table}[ht]
\centering
\caption{Missing Values Imputation Results}
\label{tab:imputation_results}
  \setlength{\extrarowheight}{2pt}
  \renewcommand{\arraystretch}{1.2}
\begin{tabular}{l|c|c|c}
\toprule
\textbf{Models Used}    & \textbf{MSE}               & \textbf{MAE}               & \textbf{RMSE}              \\
\midrule

Iterative Imputation       & 0.46214                           & 0.36617                           & 0.67981               \\

XGBoost     & 0.01447  & 0.02451                            & 0.12029    \\

SAITS                   & 0.00392         & 0.00244        & 0.091684        \\

\textbf{FAISS KNN}               & \textbf{0.00053}      & \textbf{0.00026}     & \textbf{0.02310}        \\
\bottomrule

\end{tabular}
\end{table}

\paragraph{Comparison and Selection.}
We conducted an internal evaluation of these techniques, assessing imputation performance by comparing reconstructed values against known observations in a holdout sample. Table~\ref{tab:imputation_results} summarizes the imputation accuracy and post-imputation data consistency 
for each method. Based on these findings, we selected \textbf{FAISS kNN} with \textbf{k=5} as the best-performing method. While we primarily benchmarked against standard imputation and classification approaches like kNN, XGBoost, and Transformer-based models, novel bio-inspired paradigms such as force-of-gravity-based classifiers~\cite{10.1007/978-981-15-5616-6_21} offer an emerging alternative for tasks involving spatially grounded decisions, and could be explored in future expansions of our work.

Most missing values originate from demographic attributes for certain ZIP codes. Since demographic data tends to be more correlated with geographic proximity rather than temporal trends, using kNN for imputation makes the most sense in this context. By leveraging nearest-neighbor similarity, FAISS kNN effectively reconstructs missing demographic features using information from surrounding regions. While transformer-based techniques like SAITS generally excel in time-series imputation, they underperform in this case because demographic features exhibit stronger spatial dependencies rather than temporal variations.

\subsection{Feature Selection}
\label{app:App_C9}

Although our merged dataset is richly detailed, certain attributes may be redundant or exhibit minimal correlation with accident outcomes, thereby inflating both computational costs and the risk of overfitting. To address this, we performed a systematic feature selection process using multiple techniques:

\begin{itemize}
\item 	{Random Forest Feature Importance \cite{Breiman2001}}: We trained a Random Forest model and extracted the Gini importance scores for each feature. This provided an interpretable ranking based on how much each feature contributed to reducing impurity in decision trees.
\item 	{XGBoost Feature Importance \cite{Chen_2016}}: We leveraged XGBoost’s built-in feature ranking, which assigns importance scores based on split frequency, gain, and coverage across decision trees.
\item 	{Principal Component Analysis (PCA) \cite{MACKIEWICZ1993303}}: We applied PCA to examine variance contributions across features, helping us identify and remove attributes with minimal independent information content.
\end{itemize}

Each method produced a highly similar ranking, with only minor variations (1-2 positions for some features). The next challenge was determining an optimal subset of features to retain. We employed the following experiment to select the appropriate number of features:

\paragraph{Step 1: Performance vs. Feature Count Analysis.}
To systematically determine an appropriate feature subset size, we trained a baseline accident prediction model using subsets of increasing feature counts (from the most important feature up to all 115 features). We evaluated model performance using Accuracy, F1-score, and AUC.

\paragraph{Step 2: Identification of the Performance Plateau.}
We observed that model performance increased significantly when adding the first few features, then gradually plateaued beyond a certain threshold. Specifically, we noted that beyond 21 features, the gain in predictive performance was marginal (less than a 0.5\% increase in AUC) while computational complexity continued to rise.

\paragraph{Step 3: Feature Stability Across Methods.}
We further validated our choice by analyzing the stability of selected features across Random Forest, XGBoost, and PCA. The top 21 features were consistently ranked highly across all three methods, reinforcing their importance, and beyond 21 features, we observed increased variance in validation performance, suggesting potential overfitting. Hence, by limiting the feature count to 21, we achieved a balance between model generalization and efficiency. Table~\ref{tab:selected_features} lists all the selected 21 features.

\begin{table*}[h]
\centering
\caption{Selected Features for Traffic Accident Prediction}
\label{tab:selected_features}

  \setlength{\extrarowheight}{2pt}

  \renewcommand{\arraystretch}{1.2}
\begin{tabular}{l|p{10cm}}
\toprule
\textbf{Feature Name} & \textbf{Description} \\
\midrule
Date & Captures overall time trends and seasonality. \\

Day & Identifies whether the accident occurred on a weekday or weekend. \\

Month & Helps model seasonal effects on traffic patterns. \\

relh & Relative Humidity expressed as a percentage. \\

alti & Pressure altimeter measurement in inches. \\

drct & Wind Direction in degrees, measured from true north. 
\\
tmpf & Air Temperature in Fahrenheit, typically measured at 2 meters above the ground. \\

dwpf & Dew Point Temperature in Fahrenheit, typically measured at 2 meters above the ground. \\

sknt & Wind Speed in knots. \\

Rush Hour & Indicates whether the accident occurred during peak traffic hours. \\

Season & Represents broader seasonal trends impacting road safety. \\

vsby & Visibility in miles. \\

skyc1 & Sky Level 1 Coverage. \\

Traffic\_Signal & Presence of a traffic signal, affecting vehicle stopping and interactions. \\

Part of Day & Distinguishes between morning, afternoon, evening, and night-time driving patterns. \\

p01i & One-hour precipitation amount in inches, recorded from the observation time to the previous hourly precipitation reset. This measurement may include melted frozen precipitation depending on the sensor. \\

Crossing & Indicates the presence of pedestrian crossings, increasing accident risks. \\

US Holiday & Identifies national holidays, which influence traffic patterns and congestion. \\

population\_density & Measures urbanization levels, which impact accident risk and traffic volume. \\

median\_home\_value & Socioeconomic proxy for regional traffic infrastructure and urban development. \\

housing\_occupancy & Captures transient populations, influencing local driving behavior. \\
\_renter\_occupied & \\
\bottomrule
\end{tabular}
\end{table*}

\clearpage

\end{document}